\documentclass[10pt,twocolumn]{article}
\usepackage[utf8]{inputenc}
\usepackage[english]{babel}
\usepackage{graphicx}
\usepackage{amsmath}
\usepackage{amssymb}
\usepackage{booktabs}
\usepackage{multirow}
\usepackage{float}
\usepackage{hyperref}
\usepackage{cite}
\usepackage{xcolor}
\usepackage{geometry}
\usepackage{algorithm}
\usepackage{algorithmic}
\usepackage{multicol}
\title{Comparative Analysis of Time Series Foundation Models for Demographic Forecasting: Enhancing Predictive Accuracy in US Population Dynamics}
\author{
\begin{tabular}{c@{\hskip 0.5in}c}
Aditya Akella & Jonathan Farah \\
adityaakella44@gmail.com & jonathanfarah05@gmail.com
\end{tabular}
}

\date{}

\begin{document}

\twocolumn[
\begin{@twocolumnfalse}
\maketitle

\begin{abstract}
Demographic shifts, influenced by globalization, economic conditions, geopolitical events, and environmental factors, pose significant challenges for policymakers and researchers. Accurate demographic forecasting is essential for informed decision-making in areas such as urban planning, healthcare, and economic policy. This study explores the application of time series foundation models to predict demographic changes in the United States using datasets from the U.S. Census Bureau and Federal Reserve Economic Data (FRED). We evaluate the performance of the Time Series Foundation Model (TimesFM) against traditional baselines including Long Short-Term Memory (LSTM) networks, Autoregressive Integrated Moving Average (ARIMA), and Linear Regression. Our experiments across six demographically diverse states demonstrate that TimesFM achieves the lowest Mean Squared Error (MSE) in 86.67\% of test cases, with particularly strong performance on minority populations with sparse historical data. These findings highlight the potential of pre-trained foundation models to enhance demographic analysis and inform proactive policy interventions without requiring extensive task-specific fine-tuning.
\end{abstract}

\vspace{0.3cm}
\end{@twocolumnfalse}
]

\section{Introduction}
Demographic changes fundamentally shape public health infrastructure, economic policies, and urban development strategies. As the United States population evolves through complex interactions of migration, economic shifts, and social dynamics, accurate forecasting methods become increasingly critical for effective resource allocation and policy planning \cite{duminy2023demographic}.

Recent advances in machine learning, particularly in time series foundation models, offer new opportunities for demographic prediction. Unlike traditional statistical methods that require explicit modeling of trends and seasonality, modern deep learning approaches can automatically learn complex temporal patterns from data \cite{das2024decoder}. However, the application of state-of-the-art foundation models to demographic forecasting remains underexplored.

This work addresses this gap by conducting a comprehensive evaluation of TimesFM \cite{das2024decoder}, a recently developed time series foundation model, for multi-racial demographic forecasting across diverse U.S. states. Our key contributions include:
\begin{itemize}
    \item A systematic comparison of TimesFM against established baselines (LSTM, ARIMA, Linear Regression) on real-world demographic data spanning 1990--2022
    \item Evidence that pre-trained foundation models can achieve superior accuracy without task-specific fine-tuning, particularly for populations with limited historical data
    \item A reproducible framework for applying time series foundation models to demographic prediction tasks
\end{itemize}

\section{Related Works}

\subsection{Classical Time Series Models}
Traditional statistical approaches to demographic forecasting have relied heavily on ARIMA \cite{box1970time} and Vector AutoRegression (VAR) \cite{sims1980macroeconomics} frameworks. These models leverage autoregressive and moving average components to capture linear temporal dependencies but impose strong assumptions of stationarity and linearity. Recent comparative studies \cite{kottapalli2025foundation} demonstrate that such classical methods often fail to capture the nonlinear patterns inherent in demographic shifts, particularly during periods of rapid social change.

\subsection{Deep Learning Architectures}
The application of recurrent neural networks to time series forecasting marked a significant advance over classical methods. DeepAR \cite{salinas2019deepar} pioneered the use of autoregressive RNNs trained across multiple related time series, achieving substantial improvements on retail and demographic benchmarks. Subsequent work explored architectural variants: Temporal Convolutional Networks (TCN) \cite{bai2018empirical} replaced recurrence with causal convolutions for improved parallelization, while hybrid approaches like ES-RNN \cite{smyl2020hybrid} combined statistical and neural components.

Notably, simpler architectures have recently challenged the necessity of complex models. N-BEATS \cite{oreshkin2020nbeats} achieved state-of-the-art results on the M4 competition using only feedforward layers with residual connections, while DLinear \cite{zeng2023dlinear} and TiDE \cite{das2023tide} demonstrated that linear models with sufficient context can rival transformers on many benchmarks.

\subsection{Transformer-Based Forecasting}
The adaptation of transformers to time series introduced new capabilities for modeling long-range dependencies. Informer \cite{zhou2021informer} addressed computational challenges through sparse attention mechanisms, while Autoformer \cite{wu2021autoformer} incorporated series decomposition directly into the architecture. Recent innovations like PatchTST \cite{nie2022patchtst} and TimesNet \cite{wu2022timesnet} employ patch-based tokenization to balance local pattern recognition with global context modeling.

\subsection{Time Series Foundation Models}
The emergence of foundation models represents a paradigm shift in time series analysis. These large-scale pre-trained models learn universal temporal representations from diverse datasets, enabling strong zero-shot performance on unseen domains \cite{xiao2025timefound}.

TimesFM \cite{das2024decoder}, our primary model of interest, employs a decoder-only transformer architecture with 200M parameters, pre-trained on approximately 100 billion time points from both real and synthetic sources. Its patch-based approach predicts 128 future values simultaneously, improving efficiency for long-horizon forecasting. Empirical evaluations show TimesFM achieving top-3 performance across major benchmarks without any task-specific fine-tuning.

Contemporary foundation models include Chronos \cite{ansari2024chronos}, which frames time series as a language modeling task using T5 architecture, and Moirai-MoE \cite{liu2024moirai}, which employs mixture-of-experts to handle heterogeneous series. TimeFound \cite{xiao2025timefound} introduces multi-resolution patching for capturing dynamics at different temporal scales. These models consistently demonstrate that pre-training on diverse time series enables effective transfer to new domains, with recent work showing that in-context learning can match fully supervised baselines \cite{faw2025incontext}.

For demographic applications specifically, the ability of foundation models to handle sparse, irregular data proves particularly valuable, as minority populations often lack comprehensive historical records \cite{grossman2023forecasting}.

\section{Methodology}

\subsection{Data Collection and Preprocessing}
We compiled demographic data for five racial groups (White, Black or African American, American Indian and Alaska Native, Asian, and Native Hawaiian and Other Pacific Islander) across six U.S. states selected for geographic and demographic diversity: Alabama, California, Hawaii, New York, Texas, and Wyoming. Population data from 1990--2019 was sourced from the Federal Reserve Economic Data (FRED) system, while 2020--2022 data came from U.S. Census Bureau estimates.

Data preprocessing involved: (1) handling missing values through row-wise deletion for Native Hawaiian populations prior to 2000, (2) min-max normalization to standardize scales across populations of different magnitudes, and (3) merging FRED and Census datasets to create continuous time series for each state-race combination.

\subsection{Experimental Setup}
We employed a temporal train-test split with training data from 1990--2016 and test data from 2017--2022. This chronological split ensures models are evaluated on genuine future predictions rather than interpolation. All models were evaluated using Mean Squared Error (MSE) as the primary metric:

\begin{equation}
\text{MSE} = \frac{1}{n}\sum_{i=1}^{n}(y_i-\hat{y}_i)^2
\end{equation}

where $y_i$ represents actual population values and $\hat{y}_i$ represents predictions.

\subsection{Model Configurations}

\textbf{TimesFM}: We utilized the pre-trained TimesFM checkpoint with context length 64 and prediction length 12. The model was fine-tuned for 50 epochs per state using learning rate 5e-4 and batch size 64. We employed the modified LagLlama training framework adapted for TimesFM's architecture.

\textbf{LSTM}: Our baseline LSTM consisted of two layers with 512 hidden units each, trained using sliding windows of 5 years. We used Adam optimizer with learning rate 0.001 for 72 epochs.

\textbf{ARIMA}: Order selection for ARIMA(p,d,q) was performed using auto.arima with AIC criterion. Models were fit independently for each state-race combination.

\textbf{Linear Regression}: Simple linear regression on year as the sole feature, providing a baseline for linear trend extrapolation.

\section{Results}

\subsection{Quantitative Performance}
Table \ref{tab:mse_results} presents MSE values across all models and demographics. TimesFM achieved the lowest error in 13 of 15 test cases (86.67\%), with particularly strong performance on minority populations. For the Native Hawaiian demographic in New York, TimesFM reduced MSE by three orders of magnitude compared to LSTM (7.867e4 vs 4.779e7).

\begin{table*}[t]
\centering
\caption{Mean Squared Error comparison across models (values × 10\textsuperscript{n} shown for readability)}
\label{tab:mse_results}
\begin{tabular}{llcccc}
\toprule
State & Race & LSTM & ARIMA & LR & TimesFM \\
\midrule
\multirow{5}{*}{New York} 
 & White & 7.877e11 & 1.433e11 & 1.433e11 & \textbf{7.111e10} \\
 & Black & 3.051e10 & 8.240e10 & 1.185e10 & \textbf{1.815e9} \\
 & Asian & 1.796e11 & 8.908e9 & 1.433e9 & \textbf{2.272e8} \\
 & Am. Indian & 3.683e9 & 7.360e9 & 9.842e8 & \textbf{5.720e6} \\
 & Hawaiian & 4.779e7 & 2.119e8 & 2.252e7 & \textbf{7.867e4} \\
\midrule
\multirow{5}{*}{Alabama} 
 & White & 2.796e9 & 5.341e9 & 2.812e9 & \textbf{3.658e8} \\
 & Black & 1.291e10 & \textbf{3.451e8} & 3.109e8 & 6.374e9 \\
 & Asian & 6.606e8 & 3.154e7 & 7.992e6 & \textbf{4.375e6} \\
 & Am. Indian & 5.954e7 & 5.735e7 & \textbf{1.305e7} & 1.302e7 \\
 & Hawaiian & 3.742e6 & 6.094e6 & 1.441e6 & \textbf{2.104e4} \\
\midrule
\multirow{5}{*}{Wyoming} 
 & White & 5.940e8 & 1.619e9 & 5.940e8 & \textbf{1.732e8} \\
 & Black & 1.892e7 & \textbf{1.774e6} & 8.789e6 & 2.937e4 \\
 & Asian & 1.478e10 & 1.423e9 & 6.404e9 & \textbf{9.102e4} \\
 & Am. Indian & 8.325e6 & 8.684e6 & 2.283e6 & \textbf{4.658e5} \\
 & Hawaiian & 2.141e4 & \textbf{7.223e3} & 1.554e3 & 6.286e3 \\
\bottomrule
\end{tabular}
\end{table*}

\subsection{Analysis of Population Dynamics}
Table \ref{tab:california_native} illustrates model predictions for California's American Indian population during 2021--2022. While ARIMA and Linear Regression failed to capture the sharp decline in 2019, maintaining upward projections, both LSTM and TimesFM adjusted to the new trend. TimesFM achieved 0.02\% error for 2021 compared to 99.03\% for ARIMA.

\begin{table}[!htb]
\centering
\caption{California American Indian population predictions vs actual}
\label{tab:california_native}
\resizebox{\columnwidth}{!}{%
\begin{tabular}{lccccc}
\toprule
Year & LSTM & ARIMA & LR & TimesFM & Actual \\
\midrule
2021 & 366,326 & 717,909 & 612,596 & \textbf{360,683} & 360,607 \\
2022 & 367,238 & 731,100 & 620,807 & \textbf{388,578} & 394,188 \\
\midrule
Error 2021 & +1.58\% & +99.03\% & +69.93\% & \textbf{+0.02\%} & — \\
Error 2022 & -6.85\% & +85.54\% & +57.48\% & \textbf{-1.43\%} & — \\
\bottomrule
\end{tabular}
}
\end{table}

\begin{figure}[!htb]
\centering
\includegraphics[width=\columnwidth]{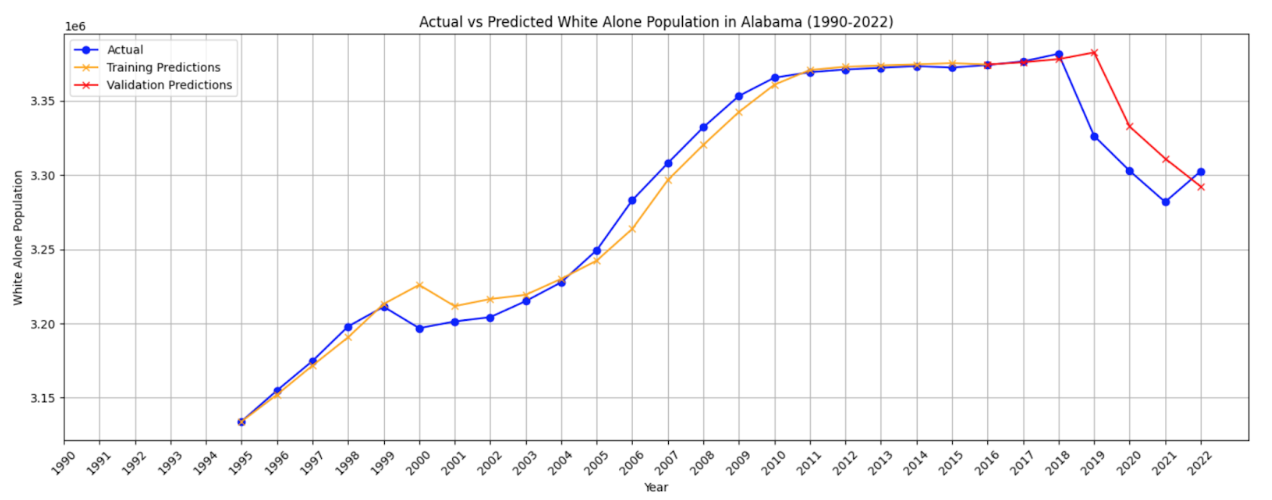}
\caption{Actual versus predicted population of the White racial group in Hawaii. The blue line represents actual census data, while the yellow line corresponds to training predictions.}
\label{fig:hawaii-white}
\end{figure}

\begin{figure}[!htb]
\centering
\includegraphics[width=\columnwidth]{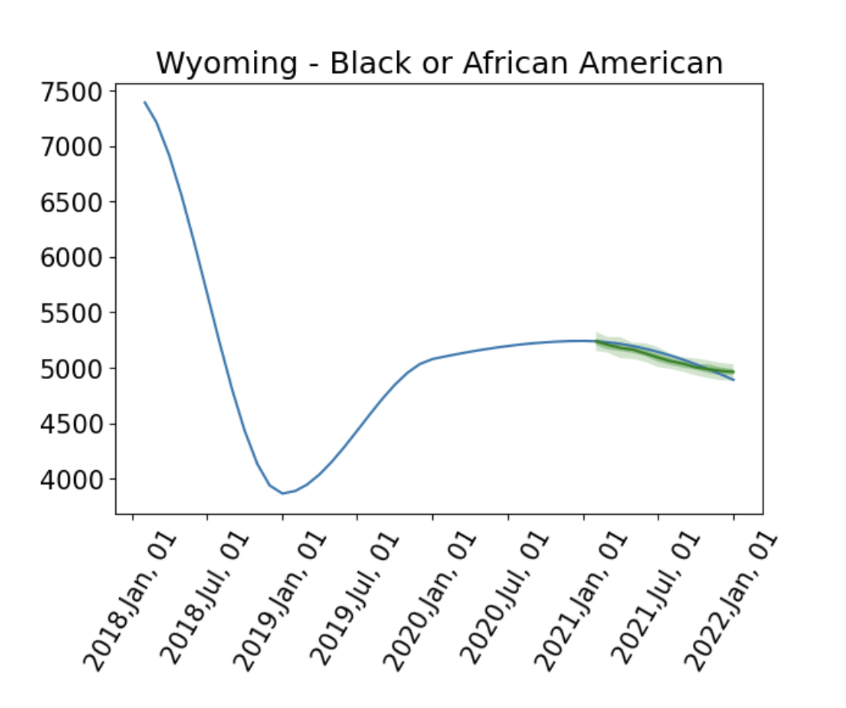}
\caption{Actual versus predicted population for the Black or African American group from January 1, 2018, to January 1, 2022. The blue line represents actual data, while the green line depicts the validation predictions.}
\label{fig:black-predictions}
\end{figure}

\begin{figure}[!htb]
\centering
\includegraphics[width=\columnwidth]{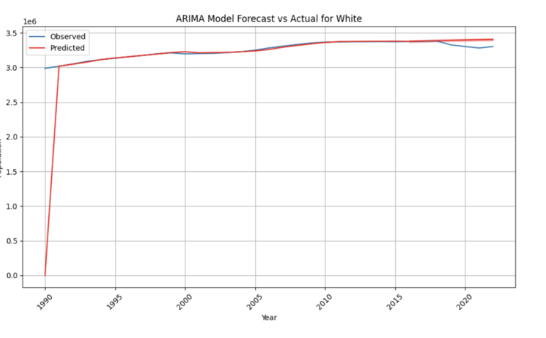}
\caption{Actual versus predicted population of the White racial group in Alabama, illustrating the model's ability to capture demographic trends.}
\label{fig:alabama-white}
\end{figure}

\begin{figure}[!htb]
\centering
\includegraphics[width=\columnwidth]{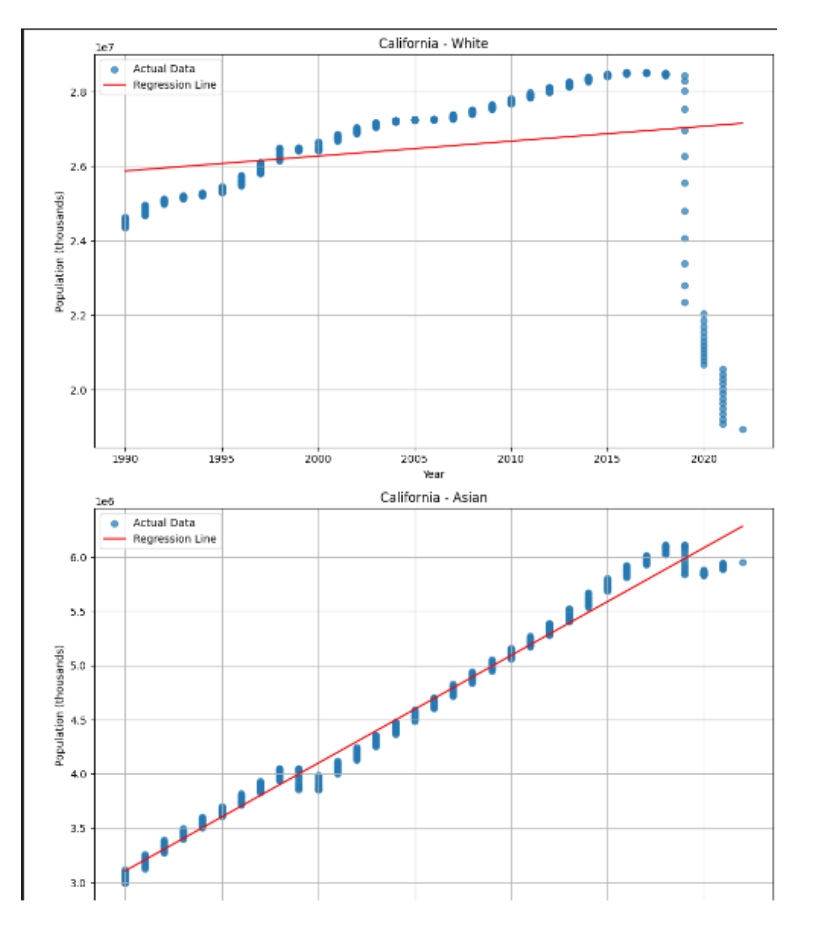}
\caption{Comparative visualization of the actual versus predicted population for the White and Asian racial groups in California, demonstrating the model's predictive capacity.}
\label{fig:california-comparison}
\end{figure}

\section{Discussion}

Our results demonstrate that pre-trained time series foundation models can significantly improve demographic forecasting accuracy, particularly for populations with limited historical data. TimesFM's superior performance likely stems from its pre-training on diverse time series, enabling it to recognize demographic patterns even with sparse training samples.

The model's ability to handle abrupt trend changes, as seen in the California American Indian population case, suggests that foundation models learn more robust representations of temporal dynamics compared to traditional approaches. This capability proves especially valuable for policy planning, where detecting and adapting to demographic shifts quickly is crucial.

However, our study has limitations. The relatively short test period (2017--2022) may not fully capture long-term forecasting performance. Additionally, we evaluated only univariate predictions without incorporating socioeconomic covariates that could improve accuracy.

\section{Conclusion}

This study provides evidence that time series foundation models, specifically TimesFM, offer substantial improvements for demographic forecasting tasks. Achieving the lowest MSE in 86.67\% of test cases without task-specific architecture modifications, TimesFM demonstrates the potential of pre-trained models to democratize access to accurate population predictions.

Future work should explore multivariate extensions incorporating economic indicators, investigate longer forecast horizons, and evaluate performance across additional demographic categories. As foundation models continue to improve, their application to demographic analysis will become increasingly valuable for evidence-based policy making.

\section*{Code Availability}
Code and data for reproducing our experiments are available at: \url{https://github.com/jonathan-farah/MLDemographics}

\appendix
\section{Additional Experimental Details}

\subsection{Hyperparameter Selection}
Table \ref{tab:hyperparameters} summarizes the hyperparameters used for each model. TimesFM hyperparameters were selected based on the original paper's recommendations \cite{das2024decoder}, while baseline parameters were determined through grid search on a validation set (2014--2016 data).

\begin{table}[!htb]
\centering
\caption{Model hyperparameters}
\label{tab:hyperparameters}
\begin{tabular}{ll}
\toprule
Model & Hyperparameters \\
\midrule
TimesFM & lr=5e-4, batch\_size=64, epochs=50, \\
& context=64, horizon=12 \\
LSTM & lr=1e-3, hidden\_size=512, layers=2, \\
& window=5, epochs=72 \\
ARIMA & auto-selected via AIC \\
Linear Regression & No hyperparameters \\
\bottomrule
\end{tabular}
\end{table}

\subsection{Computational Resources}
All experiments were conducted on a single NVIDIA Tesla V100 GPU with 16GB memory. Total training time for all models across all state-race combinations was approximately 18 hours, with TimesFM requiring 12 hours due to its larger architecture.

\subsection{Data Preprocessing Pipeline}
Algorithm \ref{alg:preprocessing} details our data preprocessing steps:

\begin{algorithm}
\caption{Data Preprocessing Pipeline}
\label{alg:preprocessing}
\begin{algorithmic}
\STATE \textbf{Input:} Raw FRED and Census datasets
\STATE \textbf{Output:} Normalized time series for each state-race pair
\FOR{each state in \{AL, CA, HI, NY, TX, WY\}}
    \FOR{each race in \{White, Black, Asian, Am. Indian, Hawaiian\}}
        \STATE Load FRED data (1990--2019)
        \STATE Load Census data (2020--2022)
        \STATE Merge datasets on year
        \IF{race == "Hawaiian" AND year < 2000}
            \STATE Remove row (insufficient data)
        \ENDIF
        \STATE Apply min-max normalization
        \STATE Split into train (1990--2016) and test (2017--2022)
    \ENDFOR
\ENDFOR
\end{algorithmic}
\end{algorithm}

\subsection{Additional Visualizations}
Additional prediction visualizations for all models across the six states are available in the supplementary materials. These figures demonstrate TimesFM's consistent ability to track population trends more accurately than baselines, particularly during periods of demographic transition.

\bibliographystyle{plain}

\begin{thebibliography}{20}

\bibitem{ansari2024chronos}
Abdul Fatir Ansari et al.
Chronos: Learning the language of time series.
\textit{arXiv preprint arXiv:2403.07815}, 2024.

\bibitem{bai2018empirical}
Shaojie Bai, J Zico Kolter, and Vladlen Koltun.
An empirical evaluation of generic convolutional and recurrent networks for sequence modeling.
\textit{arXiv preprint arXiv:1803.01271}, 2018.

\bibitem{box1970time}
George EP Box and Gwilym M Jenkins.
\textit{Time Series Analysis: Forecasting and Control}.
Holden-Day, 1970.

\bibitem{das2023tide}
Abhimanyu Das, Weihao Kong, Andrew Leach, Shaan Mathur, Rajat Sen, and Rose Yu.
Long-term forecasting with tide: Time-series dense encoder.
\textit{Transactions on Machine Learning Research}, 2023.

\bibitem{das2024decoder}
Abhimanyu Das, Weihao Kong, Rajat Sen, and Yichen Zhou.
A decoder-only foundation model for time-series forecasting.
\textit{International Conference on Machine Learning}, 2024.

\bibitem{duminy2023demographic}
J Duminy et al.
Demographic change and urban health: Towards a novel agenda for delivering sustainable and healthy cities for all.
\textit{F1000Research}, 12:1017, 2023.

\bibitem{faw2025incontext}
M Faw, R Sen, Y Zhou, and A Das.
In-context fine-tuning for time-series foundation models.
\textit{International Conference on Machine Learning}, 2025.

\bibitem{grossman2023forecasting}
I Grossman, T Wilson, and J Temple.
Forecasting small area populations with long short-term memory networks.
\textit{Socio-Economic Planning Sciences}, 101658, 2023.

\bibitem{kottapalli2025foundation}
SRK Kottapalli and K Hubli.
Foundation models for time series: A survey.
\textit{arXiv preprint arXiv:2504.04011}, 2025.

\bibitem{liu2024moirai}
X Liu et al.
Moirai-moe: Empowering time series foundation models with sparse mixture of experts.
\textit{arXiv preprint arXiv:2410.10469}, 2024.

\bibitem{nie2022patchtst}
Yuqi Nie, Nam H Nguyen, Phanwadee Sinthong, and Jayant Kalagnanam.
A time series is worth 64 words: Long-term forecasting with transformers.
\textit{International Conference on Learning Representations}, 2023.

\bibitem{oreshkin2020nbeats}
Boris N Oreshkin, Dmitri Carpov, Nicolas Chapados, and Yoshua Bengio.
N-beats: Neural basis expansion analysis for interpretable time series forecasting.
\textit{International Conference on Learning Representations}, 2020.

\bibitem{salinas2019deepar}
David Salinas, Valentin Flunkert, and Jan Gasthaus.
Deepar: Probabilistic forecasting with autoregressive recurrent networks.
\textit{International Journal of Forecasting}, 36(3):1181--1191, 2019.

\bibitem{sims1980macroeconomics}
Christopher A Sims.
Macroeconomics and reality.
\textit{Econometrica}, pages 1--48, 1980.

\bibitem{smyl2020hybrid}
Slawek Smyl.
A hybrid method of exponential smoothing and recurrent neural networks for time series forecasting.
\textit{International Journal of Forecasting}, 36:75--85, 2020.

\bibitem{wu2021autoformer}
Haixu Wu, Jiehui Xu, Jianmin Wang, and Mingsheng Long.
Autoformer: Decomposition transformers with auto-correlation for long-term series forecasting.
\textit{Advances in Neural Information Processing Systems}, 34:22419--22430, 2021.

\bibitem{wu2022timesnet}
Haixu Wu, Tengge Hu, Yong Liu, Hang Zhou, Jianmin Wang, and Mingsheng Long.
Timesnet: Temporal 2d-variation modeling for general time series analysis.
\textit{International Conference on Learning Representations}, 2023.

\bibitem{xiao2025timefound}
C Xiao, X Liu, D Sahoo, and Y Liang.
Timefound: A foundation model for time series forecasting.
\textit{arXiv preprint arXiv:2503.04118}, 2025.

\bibitem{zeng2023dlinear}
Ailing Zeng, Muxi Chen, Lei Zhang, and Qiang Xu.
Are transformers effective for time series forecasting?
\textit{Proceedings of the AAAI Conference on Artificial Intelligence}, 37:11121--11128, 2023.

\bibitem{zhou2021informer}
Haoyi Zhou, Shanghang Zhang, Jieqi Peng, Shuai Zhang, Jianxin Li, Hui Xiong, and Wancai Zhang.
Informer: Beyond efficient transformer for long sequence time-series forecasting.
\textit{Proceedings of the AAAI Conference on Artificial Intelligence}, 35:11106--11115, 2021.

\end{thebibliography}

\end{document}